\documentclass[pmlr]{jmlr}

\jmlrvolume{}
\jmlryear{}
\jmlrpages{}
\jmlrworkshop{}
\jmlrproceedings{}{}

\usepackage{fancyhdr}
\fancypagestyle{plain}{%
  \fancyhf{}                
  \cfoot{\thepage}          
}
\usepackage{longtable}

\usepackage{booktabs}
\usepackage[load-configurations=version-1]{siunitx} 

\theorembodyfont{\upshape}
\theoremheaderfont{\scshape}
\theorempostheader{:}
\theoremsep{\newline}

\title[Heart Disease Prediction]{Heart Disease Prediction: 
A Comparative Study of Optimizers' Performance in Deep Neural Networks}

\author{
  \Name{Chisom Chibuike}\thanks{This work was done while Author~1 was a research intern at SAIL Innovation Lab, supervised by Author~2.}
  \Email{chisom.chibuike.246093@unn.edu.ng} \\
  \addr{University of Nigeria}
  \AND
  \Name{Adeyinka Ogunsanya} 
  \Email{adeyinka@cchub.africa} \\
  \addr{SAIL Innovation Lab}
}

\begin{document}

\maketitle

\thispagestyle{plain}

\begin{abstract}
Optimization has been an important factor and topic of interest in training deep learning models, yet less attention has been given to how we select the optimizers we use to train these models. Hence, there is need to dive deeper into how we select the optimizers we use for training and the metrics that determine this selection. In this work, we compare the performance of 10 different optimizers in training a simple Multi-layer Perceptron model using a heart disease dataset from Kaggle. We set up a consistent training paradigm and evaluate the optimizers based on metrics such as convergence speed and stability. We also include some other Machine Learning Evaluation metrics such as AUC, Precision, and Recall, which are central metrics to classification problems. Our results show that there are trade-offs between convergence speed and stability, as optimizers like Adagrad and Adadelta, which are more stable, took longer time to converge. Across all our metrics, we choose RMSProp to be the most effective optimizer for this heart disease prediction task because it offered a balanced performance across key metrics. It achieved a precision of 0.765, recall of 0.827, and an AUC of 0.841, along with faster training time. However, it was not the most stable. We recommend that in less compute constrained environment, this method of choosing optimizers through a thorough evaluation should be adopted to increase the scientific nature and performance in training deep learning models.
\end{abstract}


\section{Introduction}
\label{sec:intro}




In recent years, Deep Neural Networks (DNNs) have been central to advances in artificial intelligence, driving progress across a wide array of applications, including autonomous systems, medical diagnostics, natural language processing, and beyond \citep{schm, unknownher}. Despite their impressive representational power and scalability, the practical effectiveness of DNNs is strongly influenced by the optimization algorithms employed during training \citep{huang2020normalizationtechniquestrainingdnns}.

Optimizers serve as the engine of the learning process, updating network parameters to minimize a given loss function. The choice of optimizer directly impacts key aspects of training; convergence speed, stability, generalization, and overall predictive performance \citep{elharrouss2025lossfunctionsdeeplearning}. While a variety of optimizers have been proposed in several literature, including both adaptive and non-adaptive methods \citep{DESOUZA2021107254}, there is limited systematic literature empirically evaluating their comparative performance in building real-world solutions.

This paper presents a comprehensive empirical study comparing the performance of ten widely used optimization algorithms: Stochastic Gradient Descent (SGD)~\citep{robbins1951stochastic, nemirovski1978cezari, shalev2007pegasos, hardt2016train}, Stochastic Gradient Descent with Nesterov Momentum (SGD-Nesterov)~\citep{pmlr-v28-sutskever13}, Root Mean Square Propagation (RMSProp)~\citep{tieleman}, Adaptive Gradient Algorithm (Adagrad)~\citep{JMLR:v12:duchi11a}, Adaptive Delta (Adadelta)~\citep{zeiler2012adadeltaadaptivelearningrate}, Adaptive Moment Estimation (Adam)~\citep{kingma2017adammethodstochasticoptimization}, Adam with Decoupled Weight Decay (AdamW)~\citep{loshchilov2019decoupledweightdecayregularization}, Infinity Norm-based Adam (Adamax)~\citep{JMLR:v12:duchi11a}, Adaptive Maximum Smoothing Gradient (AMSGrad)~\citep{reddi2019convergenceadam}, and Nesterov-accelerated Adaptive Moment Estimation (Nadam)~\citep{Dozat}.
Our evaluation is performed on a heart disease dataset from kaggle \citep{heart-dataset}, using a fully connected deep neural network architecture with six hidden layers.

Each optimizer is evaluated across multiple metrics, such as; Convergence Speed, Training Stability (measured via variance in validation loss), and those metrics critical to healthcare prediction tasks; Area Under the ROC Curve (AUC), precision and recall. By focusing on these metrics, we aim to provide a holistic view of the optimizers behavior and nuances, and use this as a guide to selecting the best-performing optimizer, with particular attention to robustness and reliability.

The remainder of this paper is organized as follows: \sectionref{sec:review} reviews related works on optimization in deep learning, highlighting existing comparative studies.
\sectionref{sec:algorithms} details the dataset, preprocessing pipeline, and the architecture of the deep neural network used. \sectionref{sec:results} presents experimental results and visualizations of the performance comparisons across optimizers. \sectionref{sec:conclude} concludes with key insights and suggestions for future work, particularly in the context of Optimizer's.

\section{Related Works}
\label{sec:review}
Despite the wide range of available optimizers, their comparative behavior in building deep learning algorithms remains underexplored. This section reviews prior evaluations and highlights existing gaps. These gaps motivate our empirical study on optimizer performance using a heart disease dataset.

Several studies have conducted empirical comparisons of optimizers across standard datasets. For instance, \citep{article}, evaluated six optimization algorithms—SGD\citep{robbins1951stochastic}, Nesterov momentum\citep{Dozat}, RMSprop\citep{tieleman}, Adam\citep{kingma2017adammethodstochasticoptimization}, Adagrad\citep{JMLR:v12:duchi11a}, and Adadelta\citep{zeiler2012adadeltaadaptivelearningrate}—on MNIST\citep{deng2012mnist}, Fashion-MNIST\citep{xiao2017fashionmnistnovelimagedataset}, CIFAR-10\citep{krizhevsky2009learning}, and CIFAR-100\citep{krizhevsky2009learning}. Their findings suggest that Adam\citep{kingma2017adammethodstochasticoptimization} achieved consistently superior test accuracy across all datasets, underscoring the advantage of adaptive methods in image classification tasks.

In a more domain-specific study, Yaqub et al. \cite{brainsci10070427} compared ten optimizers in the context of brain tumor segmentation using the BraTS2015\citep{6975210} dataset. Training a convolutional neural network (CNN) architecture on MRI data, they found that Adam\citep{kingma2017adammethodstochasticoptimization}  outperformed other optimizers, achieving a 99.2\% classification accuracy. Their results emphasized the sensitivity of medical imaging models to optimizer selection and hyperparameter configurations.

Other works  have explored the interplay of momentum and adaptive learning strategies. For example, \cite{inproceedings} studied diabetic detection via backpropagation networks, comparing gradient descent (GD)\cite{ruder2017overviewgradientdescentoptimization}, GD with momentum\citep{kingma2017adammethodstochasticoptimization}, GD with adaptive learning rate\citep{JMLR:v12:duchi11a}, and a hybrid approach. They reported that combining momentum and adaptive learning led to faster convergence.

Despite these efforts, prior studies often omit evaluating the interplay between training stability and convergence speed, which are vital in evaluating the performance of optimizers in deep learning models, for building real-world applications. Additionally, there remains a lack of focused comparative studies on optimizer behavior in tabular clinical datasets, especially using deep neural networks. Most prior work either targets convolutional architectures or standard benchmarks, leaving a gap in understanding optimizer effectiveness for structured, real-world healthcare data.




\section{Optimization Algorithms, DNN Architecture, and Datasets}
\label{sec:algorithms}

\subsection{Optimization Algorithms}
Optimization algorithms play a central role in training deep neural networks by minimizing a defined loss function through iterative parameter updates. In this work, we evaluate the following optimization algorithms/optimizers: SGD\citep{robbins1951stochastic}, SGD with Nesterov momentum\citep{sutskever}, RMSprop\citep{tieleman}, Adagrad\citep{JMLR:v12:duchi11a}, Adadelta\citep{zeiler2012adadeltaadaptivelearningrate}, Adam\citep{kingma2017adammethodstochasticoptimization}, AdamW\citep{loshchilov2019decoupledweightdecayregularization}, Adamax\citep{JMLR:v12:duchi11a}, Nadam\citep{Dozat}, and AMSGrad\citep{reddi2019convergenceadam}. \textit{For detailed pseudocode of the optimization algorithms used in this work, please refer to Appendix~\ref{apd:first}.}
    
\subsection{Deep Neural Network Architecture}

To ensure a controlled evaluation of the optimization algorithms, we designed a fixed DNN architecture that remained constant throughout the experiments carried out. The model used for the experiment comprises six fully connected hidden layers, structured in an hourglass topology, with neuron sizes \( [16, 32, 64, 32, 16, 8] \). Each hidden layer employs the ReLU (Rectified Linear Unit) activation function defined as:
\begin{equation}
\text{ReLU}(z) = \max(0, z)
\label{eq:relu}
\end{equation}
The output layer contains a single neuron activated by the sigmoid function:
\begin{equation}
\sigma(z) = \frac{1}{1 + e^{-z}}
\label{eq:sigmoid}
\end{equation}
which outputs a probability score \( \hat{y} \in [0, 1] \) indicating the likelihood of heart disease.
\paragraph{Input and Output Formalism}
Given an input feature vector \( \vec{x} \in \mathbb{R}^d \), where \( d \) is the number of clinical features, the network computes a prediction \( \hat{y} = f(\vec{x}; \theta) \) through a composition of affine transformations and nonlinear activations:
\begin{equation}
\hat{y} = \sigma \left( W^{(L)} \,\phi^{(L-1)}\!\left( \cdots \phi^{(1)}\!\left( W^{(1)} \vec{x} + \vec{b}^{(1)} \right) \cdots \right) + \vec{b}^{(L)} \right)
\label{eq:dnn_forward}
\end{equation}
Here, \( L = 7 \) which denotes the total number of layers (6 hidden + 1 output) \ \footnote{The input layer is not counted among the \( L = 7 \) layers because it does not involve any trainable parameters or transformations. The layer count \( L \) includes only those layers that apply affine transformations and nonlinear activations (i.e., hidden and output layers).}
\paragraph{Loss Function}
The model is trained using the binary cross-entropy loss function:
\begin{equation}
\mathcal{L}(\theta) = -\frac{1}{N} \sum_{i=1}^N \left[ y^{(i)} \log \hat{y}^{(i)} + (1 - y^{(i)}) \log (1 - \hat{y}^{(i)}) \right]
\label{eq:bce}
\end{equation}
\( y^{(i)} \in \{0, 1\} \) is the ground truth label for the \( i^{\text{th}} \) sample, and \( \hat{y}^{(i)} \) is the predicted probability.

\paragraph{Weight Reinitialization Protocol}
To ensure that the observed differences in performance are solely attributable to the optimizers under evaluation, we adopted a strict weight reinitialization strategy. The DNN was instantiated once and its initial parameters \( \theta_0 \) were stored. For every optimizer under evaluation, the model weights were reset to \( \theta_0 \) prior to training. Formally, for optimizer \( \mathcal{O}_j \), training commenced from:
\begin{equation}
\theta^{(0)}_j = \theta_0
\label{eq:init}
\end{equation}
This guarantees that each optimizer begins from an identical initialization state, thus isolating the optimizer’s parameters effect on training dynamics.

\paragraph{Initial Training Setup}
All models were trained for up to 50 epochs. Metrics such as AUC, precision, recall, convergence speed and stability (defined as the standard deviation of validation loss) were recorded per epoch. No hyperparameter tuning was performed at the initial stage.

\paragraph{Enhanced Training Setup}  
The initial training setup which was first carried-out and the best-performing optimizer was selected for further refinement. In the enhanced training phase, the initial model architecture was trained using the best-performing optimizer and augmented with dropout layers to mitigate overfitting. A dropout rate of \( p = 0.2 \) was applied after each hidden layer. This stage also incorporated early stopping with a patience of 15 epochs, along with a grid search over learning rates \( \eta \in \{0.001, 0.01, 0.1\} \) to identify the optimal learning rate. Additionally, a 5-fold cross-validation was employed to ensure robust evaluation and improve the final model's performance.\footnote{The training setup was done to guarantee a principled and reproducible framework for evaluating optimizer performance on a real-world classification task.}


\subsection{The Dataset}
\label{sec:dataset}

The dataset used in this study is sourced from a publicly available Kaggle repository\footnote{\url{https://www.kaggle.com/datasets/sid321axn/heart-statlog-cleveland-hungary-final/data}} and aggregates patient records relevant to cardiovascular health. It comprises \( N = 1190 \) samples and \( d = 11 \) features, each capturing demographic and clinical information known to correlate with heart disease. The prediction target is a binary label \( y \in \{0, 1\} \), where \( y = 1 \) indicates the presence of heart disease, and \( y = 0 \), its absence.

\begin{table}[t]  
\floatconts
  {tab:feature-summary}
  {\caption{Overview of the features in the heart disease dataset.}}
  {\scriptsize
    \begin{tabular}{|l|l|}
    \hline
    \textbf{Feature Name} & \textbf{Description} \\\hline
    \texttt{age} & Age of the patient (in years) \\
    \texttt{sex} & Gender of the patient (1 = Male, 0 = Female) \\
    \texttt{chest pain type} & Type of chest pain experienced \\
    \texttt{resting bp s} & Resting blood pressure (mm Hg) \\
    \texttt{cholesterol} & Serum cholesterol level (mg/dL) \\
    \texttt{fasting blood sugar} & Fasting blood sugar $>$ 120 mg/dL (1 = True, 0 = False) \\
    \texttt{resting ecg} & Resting electrocardiographic results \\
    \texttt{max heart rate} & Maximum heart rate achieved \\
    \texttt{exercise angina} & Exercise-induced angina (1 = Yes, 0 = No) \\
    \texttt{oldpeak} & ST depression induced by exercise relative to rest \\
    \texttt{ST slope} & Slope of the peak exercise ST segment \\
    \texttt{target} & Presence of heart disease (1 = disease, 0 = normal) \\\hline
    \end{tabular}
  }
\end{table}

\vspace{0.2cm}
\noindent\textbf{Exploratory Data Analysis (EDA).}
EDA revealed that the target variable is relatively balanced, with no significant skew toward either class, eliminating the need for resampling or class weighting. Figure~\ref{fig:target-dist} illustrates the distribution of the target labels.

\vspace{0.2cm}
\noindent\textbf{Preprocessing Pipeline.}
To ensure data consistency and prepare for model training, the following preprocessing steps were applied:
\begin{enumerate}
    \item \textbf{Missing Values and Redundancies:} The dataset contained no missing entries. However, one feature—\texttt{fasting blood sugar} was dropped due to low variance (\( >75\% \) of samples held a value of 0). Additionally, 272 duplicate rows were removed.
    \item \textbf{Invalid Zero Entries:} Certain physiological measurements (e.g., \texttt{cholesterol} and \texttt{resting bp s}) contained zero values, which are not clinically plausible. These were replaced with the feature-wise mean.
    \item \textbf{Normalization:} To reduce sensitivity to outliers, we applied Robust Scaling to all numerical features.
    \item \textbf{Feature Encoding:} All categorical features were already numerically encoded. Manual verification confirmed datatype consistency and appropriate class ranges. No additional one-hot encoding was required.
    \item \textbf{Train-Test Split:} The dataset was partitioned into training and testing sets using a 70:30 stratified split to preserve class proportions. Within the training set, 20\% of data was held out as a validation set during training.
\end{enumerate}
These processes were carried out in an attempt to eradicates any bias from the data inorder to ensure that the model is trained on clean, normalized, and representative data while avoiding information leakage or feature imbalance. The resulting dataset provides a reliable foundation for the empirical comparison of optimization algorithms.

\begin{figure}[htbp]
\floatconts
  {fig:target-dist}
  {\caption{Distribution of the binary target variable (0 = no disease, 1 = disease).}}
  {\includegraphics[width=0.5\linewidth]{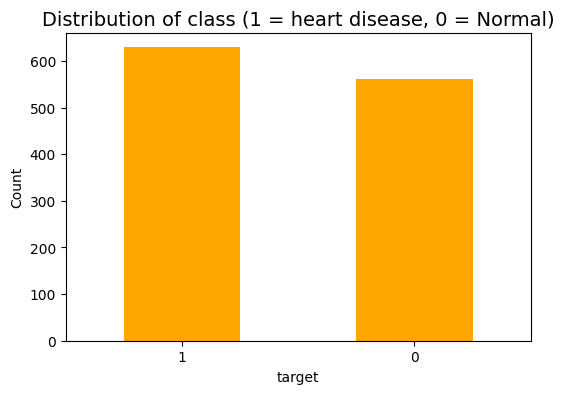}}
\end{figure}

\section{Experimental Results}
\label{sec:results}

The evaluation and results focuses on five key metrics: convergence speed, stability, precision, recall, and the Area Under the ROC Curve (AUC-ROC). These metrics are selected to capture both optimizer's dynamics and clinical relevance.

\subsubsection{Convergence Speed}
Convergence speed refers to the number of training epochs required for the model's loss function to stop improving significantly. It indicates how quickly an optimizer leads the model to reach its goal of attaining a minimum of the loss landscape, and is typically measured by the epoch at which validation loss plateaus.

To evaluate convergence speed, we analyzed how quickly each optimizer reduced the training loss to a stable minimum. Specifically, we recorded the number of epochs required for the validation loss to reach its lowest point, referred to as the \emph{convergence epoch}. The results are summarized in Table~\ref{tab:convergence-performance}, which lists the final training and validation losses along with the corresponding convergence epoch for each optimizer.

As illustrated in Figure~\ref{fig:convergence-speed}, the optimizers exhibited markedly different convergence patterns. \textbf{Adam}\citep{kingma2017adammethodstochasticoptimization}, \textbf{Nadam} \citep{Dozat}, \textbf{RMSProp} \citep{tieleman}, \textbf{AMSgrad}\citep{reddi2019convergenceadam}, and \textbf{AdamW} \citep{loshchilov2019decoupledweightdecayregularization} demonstrated rapid convergence within the first 10--18 epochs. However, this speed came at the cost of generalization performance, as we observe notable difference between their training and validation losses, suggesting overfitting. We observed that adaptive optimizers such as Adam showed stronger overfitting compared to SGD. This is consistent with prior findings that while adaptive methods improve optimization speed, they reduce implicit regularization, leading to poorer generalization~\citep{wilson2017marginal, keskar2017improving}.

\begin{table}[htbp]
\floatconts
  {tab:convergence-performance}
  {\caption{Final Training and Validation Loss and Convergence Epoch for Each Optimizer}}
  {\scriptsize
    \begin{tabular}{lccc}
    \toprule
    \bfseries Optimizer & \bfseries Final Training Loss & \bfseries Final Validation Loss & \bfseries Convergence Epoch \\
    \midrule
    SGD & 0.421812 & 0.529095 & 49 \\
    Adam & 0.114762 & 1.331397 & 9 \\
    RMSprop & 0.157385 & 0.840259 & 18 \\
    Adagrad & 0.662042 & 0.681975 & 49 \\
    Adadelta & 0.697742 & 0.701515 & 49 \\
    Adamax & 0.305551 & 0.527479 & 45 \\
    Nadam & 0.083017 & 1.239177 & 15 \\
    AMSgrad & 0.125040 & 1.127012 & 10 \\
    AdamW & 0.114762 & 1.331397 & 9 \\
    SGD Nesterov & 0.421812 & 0.529095 & 49 \\
    \bottomrule
    \end{tabular}
  }
\end{table}

\begin{figure}[t]
\floatconts
  {fig:convergence-speed}
  {\caption{Convergence curves (training and validation loss) for all optimizers over 50 epochs.}}
  {\includegraphics[scale=0.72]{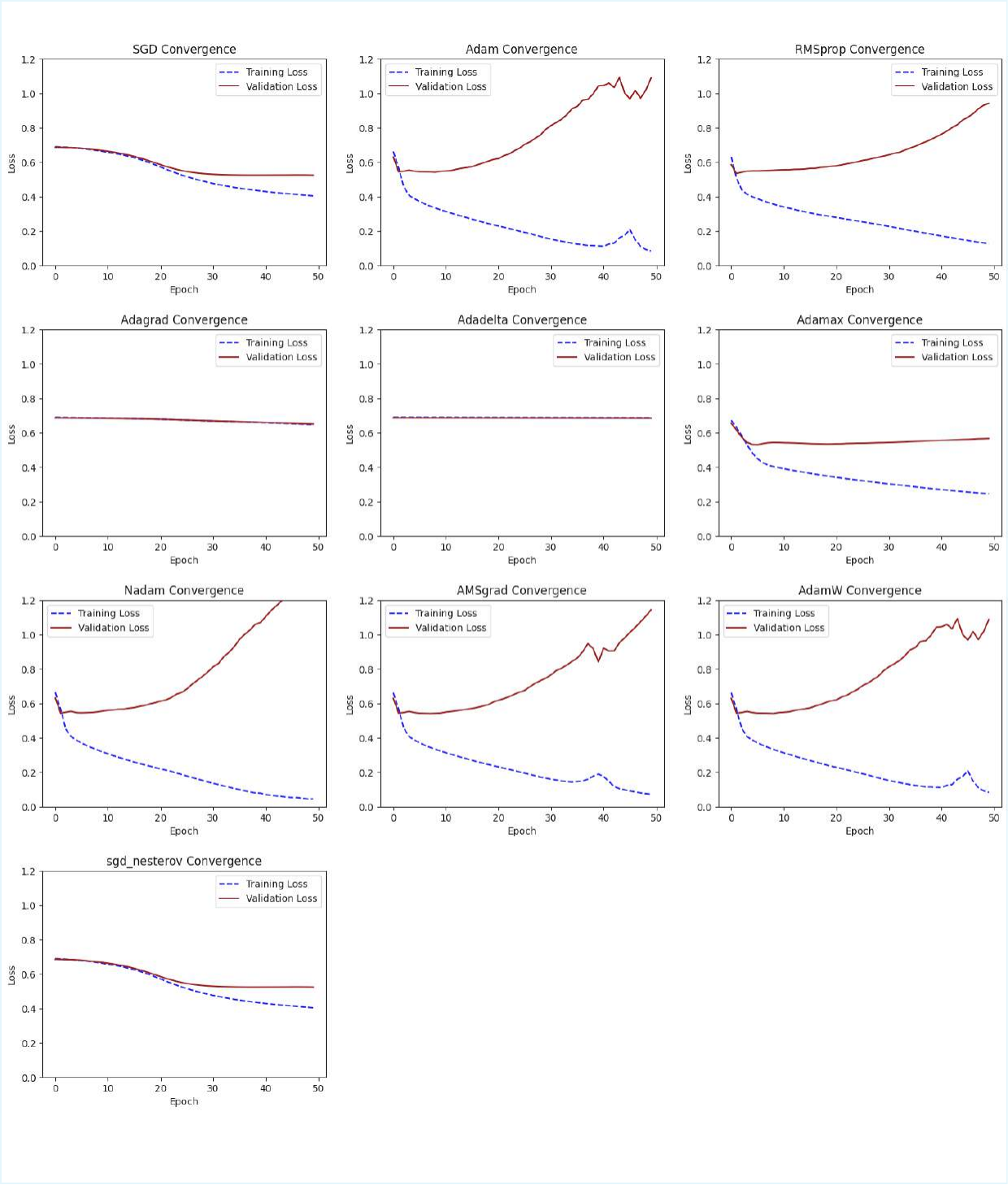}}
\end{figure}

\subsubsection{Stability.}
In this work, we refer to stability, as the standard deviation of the validation loss across training epochs. Optimizers that produce large fluctuations or oscillations in the loss trajectory are deemed less stable, which can adversely affect model reliability and convergence.

To evaluate the stability of each optimizer, we analyzed the variance in validation loss over training epochs. Optimizers that yield minimal fluctuations in validation loss are considered more stable, as they ensure smoother and more predictable learning. Table~\ref{tab:stability} presents the standard deviation of the validation loss for each optimizer, serving as a direct measure of training consistency.

\begin{table}[htbp]
\floatconts
  {tab:stability}
  {\caption{Stability of Each Optimizer Measured by Standard Deviation of Validation Loss}}%
  {\scriptsize
    \begin{tabular}{lccc}
    \toprule
    \bfseries Optimizer & \bfseries Stability (Validation Loss Std Dev) \\
    \midrule
    SGD & 0.056739 \\
    Adam & 0.214864 \\
    RMSprop & 0.094638 \\
    Adagrad & 0.005564 \\
    Adadelta & 0.001403 \\
    Adamax & 0.047790 \\
    Nadam & 0.208663 \\
    AMSgrad & 0.182095 \\
    AdamW & 0.214864 \\
    SGD Nesterov & 0.056739 \\
    \bottomrule
    \end{tabular}
}
\end{table}

The results in figure \ref{fig:stability} and table \ref{tab:stability}, indicate that; \textbf{Adadelta}\citep{zeiler2012adadeltaadaptivelearningrate}, \textbf{Adagrad}\citep{JMLR:v12:duchi11a}, and \textbf{Adamax}\citep{JMLR:v12:duchi11a} demonstrate the highest stability, exhibiting extremely low standard deviations. These optimizers provide consistent training dynamics, making their result more reliable. 

\vspace{-0.2cm}
\begin{figure}[t]
\floatconts
  {fig:stability}
  {\caption{Standard deviation of validation loss across training epochs for each optimizer.}}
  {\includegraphics[width=\linewidth,height=0.3\textheight,keepaspectratio]{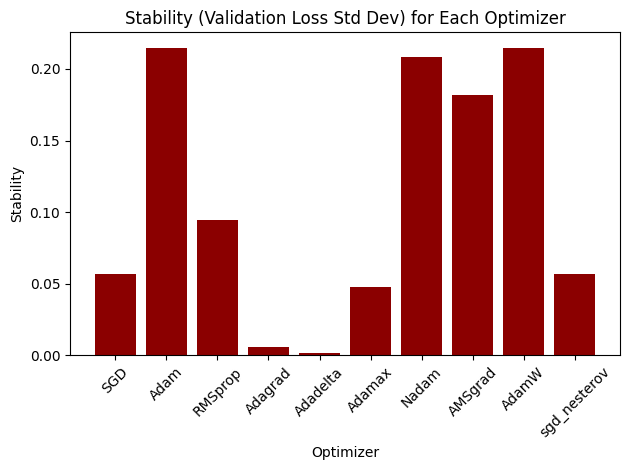}}
\end{figure}

\subsubsection{Precision, Recall, and AUC}
To assess classification and predictive performance, we evaluated each optimizer using Precision, Recall, and AUC.
Table~\ref{tab:metrics-performance} summarizes the final precision, recall, and AUC scores for each optimizer.

\begin{table}[htbp]
\floatconts
  {tab:metrics-performance}
  {\caption{Final Precision, Recall, and AUC Scores for Each Optimizer}}%
  {\scriptsize
    \begin{tabular}{lccc}
    \toprule
  \bfseries Optimizer & \bfseries Precision & \bfseries Recall & \bfseries AUC \\
  \midrule
  SGD & 0.706 & 0.800 & 0.821 \\
  Adam & 0.707 & 0.867 & 0.822 \\
  RMSprop & \textbf{0.765} & 0.827 & 0.841 \\
  Adagrad & 0.449 & 1.000 & 0.782 \\
  Adadelta & 0.449 & 1.000 & 0.385 \\
  Adamax & 0.741 & 0.800 & \textbf{0.860} \\
  Nadam & 0.731 & 0.760 & 0.843 \\
  AMSgrad & 0.756 & 0.787 & 0.819 \\
  AdamW & 0.707 & 0.867 & 0.822 \\
  SGD Nesterov & 0.706 & 0.800 & 0.821 \\
  \bottomrule
  \end{tabular}}
\end{table}

The results indicate several trade-offs among the optimizers across the evaluated metrics:

\begin{itemize}
  \item \textbf{Precision}: \textbf{RMSprop}\citep{tieleman} achieved the highest precision (0.765), followed by \textbf{AMSgrad}\citep{reddi2019convergenceadam} (0.756) and \textbf{Adamax}\citep{JMLR:v12:duchi11a} (0.741). High precision indicates that these optimizers were more effective at minimizing false positives—crucial in clinical settings to avoid unnecessary treatment for patients who are not actually at risk.

  \item \textbf{Recall}: \textbf{Adagrad}\citep{JMLR:v12:duchi11a} and \textbf{Adadelta}\citep{zeiler2012adadeltaadaptivelearningrate} achieved perfect recall (1.000), meaning they correctly identified all true positive cases. However, their low precision (0.449) implies a high rate of false positives. Furthermore, \textbf{Adam}\citep{kingma2017adammethodstochasticoptimization} and \textbf{AdamW}\citep{loshchilov2019decoupledweightdecayregularization} achieved strong recall (0.867), followed by \textbf{RMSProp}\citep{tieleman} (0.827) while maintaining higher precision than Adagrad\citep{JMLR:v12:duchi11a} and Adadelta\citep{zeiler2012adadeltaadaptivelearningrate}, thus offering a more balanced performance.

  \item \textbf{AUC}: \textbf{Adamax}\citep{JMLR:v12:duchi11a} recorded the highest AUC (0.860), indicating the best overall ability to discriminate between classes across thresholds. \textbf{Nadam}\citep{Dozat} (0.843) and \textbf{RMSprop}\citep{tieleman} (0.841) followed closely, suggesting their strong generalization and ranking performance.
\end{itemize}

Based on the comprehensive evaluation across all metrics, \textbf{RMSprop}\citep{tieleman} is identified as the most effective optimizer for the heart disease prediction task. RMSprop\citep{tieleman} demonstrates the highest final precision (0.765), a strong AUC score (0.841), and a high recall (0.827), indicating its robust ability to distinguish between positive and negative cases while minimizing false positives and false negatives. Furthermore, RMSprop\citep{tieleman} converges faster, within 18 epochs. Although Adamax\citep{JMLR:v12:duchi11a} achieves the highest AUC (0.860), its recall is comparatively lower (0.800), and its convergence is slower (45 epochs).  Overall, RMSprop\citep{tieleman} offers the best balance across predictive accuracy, generalization, convergence, efficiency, and reliability, making it the most suitable choice for this medical prediction task.

In conclusion, we trained our final model with RMSProp\citep{tieleman} and enhanced it with \textbf{dropout regularization}, \textbf{hyperparameter tuning}, and \textbf{early stopping}, achieving a robust \textbf{ROC-AUC score of 92\% from the inital 84\%}. These results highlight RMSprop’s\citep{tieleman} as a more suitable optimizer to build a reliable model for this predictive task.

\begin{figure}
\floatconts
  {fig:final-auc}
  {\caption{Final AUC of the overall model trained using RMSProp as the best-performing optimizer.}}
  {\includegraphics[width=\linewidth,height=0.3\textheight, ]{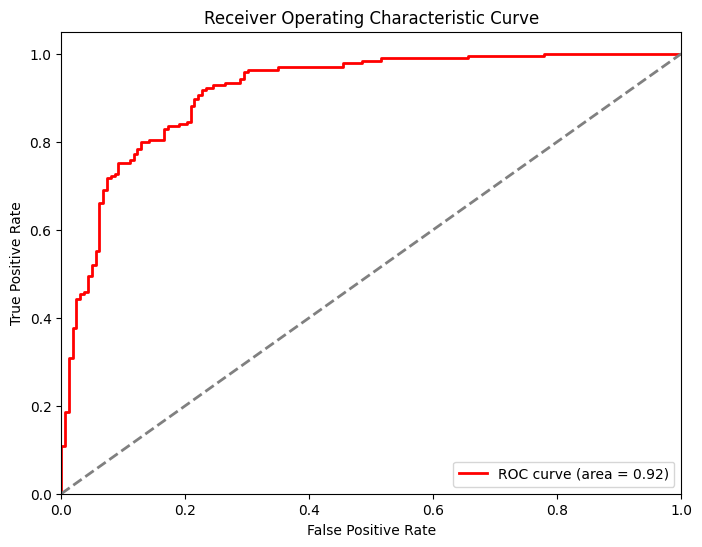}}
\end{figure}

\section{Conclusion and Discussion}
\label{sec:conclude}

This study evaluated the impact of various optimization algorithms on the performance of deep neural networks for heart disease prediction. Our findings revealed that; \textbf{RMSprop}\citep{tieleman} is the most effective optimizer for this heart disease prediction task, offering a balanced performance across key metrics. It achieved strong precision (0.765), high recall (0.827), and a solid AUC (0.841), along with faster training time. However, it is not very stable.
In addition, we noticed the trade-off between convergence speed and stability, as optimizers like Adagrad\citep{JMLR:v12:duchi11a} and Adadelta\citep{zeiler2012adadeltaadaptivelearningrate} which are more stable, took longer time to converge. By incorporating dropout regularization, hyperparameter tuning, the best optimizer for this task, and early stopping, we further enhanced the model’s ability to generalize, achieving a final ROC-AUC score of 92

These results emphasize the role of optimizers in shaping the outcome of deep learning models. Consequently, we recommend that researchers adopt a deliberate and systematic approach to evaluating optimization algorithms when designing predictive systems. Optimizers are not merely plug-and-play components as they substantially influence model training dynamics, convergence speed, and overall predictive performance. A careful selection informed by empirical evidence can lead to more accurate and robust models.

Nonetheless, our study is not without limitations. The dataset used (its size and structure) and the model's architecture, may have influenced the performance outcomes of the optimizers evaluated. Future work should investigate the generalizability of these findings by conducting experiments on larger datasets and using different learning architectures, including simpler traditional algorithms which may perform well on datasets as small as the one used in this study. Such efforts will help in establishing more reliable benchmarks and uncovering optimizer behaviors that may only emerge at scale. Further more, future work should focus on investigating the reason behind the trade-offs between the convergence speed and stability of the optimizers used in this study.

Finally, we emphasize the need for the research community to continue developing optimization techniques that are more stable, adaptive, and capable of handling the increasing complexity and diversity of real-world data. Since no single optimizer currently performs optimally across all tasks and datasets, advancing the state of optimization remains a fundamental challenge in machine learning research.

\subsection{Ethical Consideration}
The dataset used in this study is a publicly available health dataset obtained from Kaggle repository\footnote{\url{https://www.kaggle.com/datasets/sid321axn/heart-statlog-cleveland-hungary-final/data}} and was also published by \cite{dz4t-cm36-20}. It was compiled by merging multiple pre-existing datasets, including the Cleveland \citep{ritwikb3_heartdisease} heart disease datasets. According to the dataset description, all personally identifiable information has been removed, and the data has been anonymized. To the best of our knowledge, the dataset does not contain any sensitive personal health information, and its use complies with ethical standards for secondary health data.

\newpage
\appendix
\section{Algorithms}\label{apd:first}
\begin{algorithm}[http]
\floatconts
{alg:sgd}
{\caption{Stochastic Gradient Descent (SGD)}}
{
\begin{enumerate}
  \item Initialize parameters \( \theta_0 \), learning rate \( \eta \), mini-batch size \( b \), and set iteration \( t \leftarrow 0 \).
  \item Repeat until convergence:
  \begin{enumerate}
    \item Sample a mini-batch \( B_t = \{(x^{(i)}, y^{(i)})\}_{i=1}^b \) from the training set.
    \item Compute the gradient: \( g_t = \nabla_\theta J(\theta_t; B_t) \).
    \item Update the parameters: \( \theta_{t+1} = \theta_t - \eta \cdot g_t \).
    \item Increment \( t \leftarrow t + 1 \).
  \end{enumerate}
  \item Return final parameters \( \theta_t \).
\end{enumerate}
}
\end{algorithm}

\begin{algorithm}[http]
\floatconts
  {alg:adamw}
  {\caption{AdamW Optimization Algorithm \cite{loshchilov2019decoupledweightdecayregularization}}}
  {
  \begin{enumerate}
    \item Initialize parameters $\theta_0$, first moment vector $m_0 \gets 0$, 
          second moment vector $v_0 \gets 0$, timestep $t \gets 0$.
    \item Repeat until convergence:
    \begin{enumerate}
      \item Increment timestep: $t \gets t + 1$
      \item Compute gradient: $g_t = \nabla_\theta J(\theta_{t-1}; B_t)$
      \item Update biased first moment: $m_t = \beta_1 m_{t-1} + (1-\beta_1) g_t$
      \item Update biased second moment: $v_t = \beta_2 v_{t-1} + (1-\beta_2) g_t^2$
      \item Apply bias correction: 
            $\hat{m}_t = \dfrac{m_t}{1-\beta_1^t}, \;
             \hat{v}_t = \dfrac{v_t}{1-\beta_2^t}$
      \item Update parameters with weight decay:  
            $\theta_t = \theta_{t-1} - \alpha \!\left(
              \dfrac{\hat{m}_t}{\sqrt{\hat{v}_t} + \epsilon} + \lambda \theta_{t-1}
            \right)$
    \end{enumerate}
    \item Return final parameters $\theta_t$.
  \end{enumerate}
  }
\end{algorithm}

\begin{algorithm}[http]
\floatconts
{alg:adam}
{\caption{Adam \cite{Kingma&Jimmy}}}
{
\begin{enumerate}
  \item Initialize parameters \( \theta_0 \), first moment vector \( m_0 \gets 0 \), second moment vector \( v_0 \gets 0 \), timestep \( t \gets 0 \).
  \item Repeat until convergence:
  \begin{enumerate}
    \item Increment time step: \( t \gets t + 1 \)
    \item Compute gradient: \( g_t = \nabla_\theta J(\theta_{t-1}; B_t) \)
    \item Update biased first moment estimate: \( m_t = \beta_1 m_{t-1} + (1 - \beta_1) g_t \)
    \item Update biased second moment estimate: \( v_t = \beta_2 v_{t-1} + (1 - \beta_2) g_t^2 \)
    \item Bias correction:
    \begin{itemize}
        \item \( \hat{m}_t = \frac{m_t}{1 - \beta_1^t} \)
        \item \( \hat{v}_t = \frac{v_t}{1 - \beta_2^t} \)
    \end{itemize}
    \item Update parameters:\[
    \theta_t = \theta_{t-1} - \alpha \cdot \frac{\hat{m}_t}{\sqrt{\hat{v}_t} + \epsilon}
    \]
  \end{enumerate}
  \item Return final parameters \( \theta_t \).
\end{enumerate}
}
\end{algorithm}

\begin{algorithm}[http]
\floatconts
{alg:adagrad}
{\caption{Adagrad Optimization Algorithm \cite{Elad}}}
{
\begin{enumerate}
  \item Initialize parameters \( \theta_0 \), accumulator \( r_0 \gets 0 \), timestep \( t \gets 0 \).
  \item Repeat until convergence:
  \begin{enumerate}
    \item Increment time step: \( t \gets t + 1 \)
    \item Compute gradient: \( g_t = \nabla_\theta J(\theta_{t-1}; B_t) \)
    \item Update accumulator: \( r_t = r_{t-1} + g_t^2 \)
    \item Parameter update:
    \[
    \theta_t = \theta_{t-1} - \frac{\eta}{\sqrt{r_t} + \epsilon} \cdot g_t
    \]
  \end{enumerate}
  \item Return final parameters \( \theta_t \)
\end{enumerate}
}
\end{algorithm}

\begin{algorithm}[http]
\floatconts
{alg:adamax}
{\caption{Adamax Optimization Algorithm \cite{Kingma&Jimmy}}}
{
\begin{enumerate}
  \item Initialize parameters \( \theta_0 \), first moment vector \( m_0 \gets 0 \), infinity norm vector \( u_0 \gets 0 \), and timestep \( t \gets 0 \).
  \item Repeat until convergence:
  \begin{enumerate}
    \item Increment time step: \( t \gets t + 1 \)
    \item Compute gradient: \( g_t = \nabla_\theta J(\theta_{t-1}; B_t) \)
    \item Update first moment estimate: \( m_t = \beta_1 m_{t-1} + (1 - \beta_1) g_t \)
    \item Update infinity norm estimate: \( u_t = \max(\beta_2 u_{t-1}, |g_t|) \)
    \item Compute update: \( \Delta \theta_t = - \frac{\eta}{u_t + \epsilon} m_t \)
    \item Update parameters: \( \theta_t = \theta_{t-1} + \Delta \theta_t \)
  \end{enumerate}
  \item Return final parameters \( \theta_t \)
\end{enumerate}
}
\end{algorithm}

\begin{algorithm}[http]
\floatconts
{alg:rmsprop}
{\caption{RMSprop Optimization Algorithm \cite{tieleman}}}
{
\begin{enumerate}
  \item Initialize parameters \( \theta_0 \), squared gradient accumulator \( r_0 \gets 0 \), and timestep \( t \gets 0 \).
  \item Repeat until convergence:
  \begin{enumerate}
    \item Increment time step: \( t \gets t + 1 \)
    \item Compute gradient: \( g_t = \nabla_\theta J(\theta_{t-1}; B_t) \)
    \item Update exponential moving average of squared gradients:
    \[
    r_t = \beta r_{t-1} + (1 - \beta) g_t^2
    \]
    \item Update parameters:
    \[
    \theta_t = \theta_{t-1} - \frac{\eta}{\sqrt{r_t} + \epsilon} \cdot g_t
    \]
  \end{enumerate}
  \item Return final parameters \( \theta_t \)
\end{enumerate}
}
\end{algorithm}

\begin{algorithm}[http]
\floatconts
{alg:amsgrad}
{\caption{AMSGrad Optimization Algorithm \cite{Reddi}}}
{
\begin{enumerate}
  \item Initialize parameters \( \theta_0 \), first moment vector \( m_0 \gets 0 \), second moment vector \( v_0 \gets 0 \), maximum second moment \( \hat{v}_0 \gets 0 \), timestep \( t \gets 0 \).
  \item Repeat until convergence:
  \begin{enumerate}
    \item Increment time step: \( t \gets t + 1 \)
    \item Compute gradient: \( g_t = \nabla_\theta J(\theta_{t-1}; B_t) \)
    \item Update first moment estimate: \( m_t = \beta_1 m_{t-1} + (1 - \beta_1) g_t \)
    \item Update second moment estimate: \( v_t = \beta_2 v_{t-1} + (1 - \beta_2) g_t^2 \)
    \item Update maximum second moment: \( \hat{v}_t = \max(\hat{v}_{t-1}, v_t) \)
    \item Parameter update:
    \[
    \theta_t = \theta_{t-1} - \frac{\eta}{\sqrt{\hat{v}_t} + \epsilon} \cdot m_t
    \]
  \end{enumerate}
  \item Return final parameters \( \theta_t \)
\end{enumerate}
}
\end{algorithm}

\begin{algorithm}[http]
\floatconts
{alg:nesterov}
{\caption{SGD with Nesterov Momentum \cite{sutskever}}}
{
\begin{enumerate}
  \item Initialize parameters \( \theta_0 \), learning rate \( \eta \), momentum coefficient \( \mu \in [0, 1) \), and velocity vector \( v_0 \leftarrow 0 \).
  \item Set iteration \( t \leftarrow 0 \).
  \item Repeat until convergence:
  \begin{enumerate}
    \item Compute look-ahead position: \( \tilde{\theta}_t = \theta_t - \mu v_t \)
    \item Evaluate gradient at look-ahead: \( g_t = \nabla_\theta J(\tilde{\theta}_t; B_t) \)
    \item Update momentum: \( v_{t+1} = \mu v_t + \eta g_t \)
    \item Update parameters: \( \theta_{t+1} = \theta_t - v_{t+1} \)
    \item Increment \( t \leftarrow t + 1 \)
  \end{enumerate}
  \item Return final parameters \( \theta_t \)
\end{enumerate}
}
\end{algorithm}

\begin{algorithm}[http]
\floatconts
  {alg:adadelta}
  {\caption{Adadelta Optimization Algorithm \cite{Zeiler}}}
  {
  \begin{enumerate}
    \item Initialize parameters $\theta_0$, accumulated squared gradient 
          $E[g^2]_0 \gets 0$, accumulated squared updates 
          $E[\Delta \theta^2]_0 \gets 0$, and timestep $t \gets 0$.
    \item Repeat until convergence:
    \begin{enumerate}
      \item Increment timestep: $t \gets t + 1$
      \item Compute gradient: $g_t = \nabla_\theta J(\theta_{t-1}; B_t)$
      \item Update accumulated squared gradients:  
            $E[g^2]_t = \rho E[g^2]_{t-1} + (1-\rho) g_t^2$
      \item Compute parameter update:  
            $\Delta \theta_t = - \dfrac{\sqrt{E[\Delta \theta^2]_{t-1} + \epsilon}}
            {\sqrt{E[g^2]_t + \epsilon}} \, g_t$
      \item Update accumulated squared updates:  
            $E[\Delta \theta^2]_t = \rho E[\Delta \theta^2]_{t-1} + (1-\rho)\Delta\theta_t^2$
      \item Update parameters: $\theta_t = \theta_{t-1} + \Delta \theta_t$
    \end{enumerate}
    \item Return final parameters $\theta_t$
  \end{enumerate}
  }
\end{algorithm}

\begin{algorithm}[http]
\floatconts
  {alg:nadam}
  {\caption{Nadam Optimization Algorithm \cite{Dozat}}}
  {
  \begin{enumerate}
    \item Initialize parameters $\theta_0$, first moment vector $m_0 \gets 0$, 
          second moment vector $v_0 \gets 0$, timestep $t \gets 0$.
    \item Set hyperparameters: learning rate $\eta$, decay rates 
          $\beta_1, \beta_2 \in [0,1)$, and small constant $\epsilon$.
    \item Repeat until convergence:
    \begin{enumerate}
      \item Increment timestep: $t \gets t + 1$
      \item Compute gradient: $g_t = \nabla_\theta J(\theta_{t-1}; B_t)$
      \item Update biased first moment estimate:  
            $m_t = \beta_1 m_{t-1} + (1-\beta_1) g_t$
      \item Update biased second moment estimate:  
            $v_t = \beta_2 v_{t-1} + (1-\beta_2) g_t^2$
      \item Compute Nesterov-accelerated update:  
            $\tilde{g}_t = \dfrac{\eta}{\sqrt{v_t + \epsilon}} 
            \left( \beta_1 m_t + \dfrac{(1-\beta_1) g_t}{1 - \beta_1^t} \right)$
      \item Update parameters: $\theta_t = \theta_{t-1} - \tilde{g}_t$
    \end{enumerate}
    \item Return final parameters $\theta_t$
  \end{enumerate}
  }
\end{algorithm}
\end{document}